\pdfoutput=1

\documentclass[11pt]{article}

\usepackage{ACL2023}

\usepackage{multirow}
\usepackage{amsmath}
\usepackage{geometry}
\usepackage{booktabs}
\usepackage{amssymb}
\usepackage{colortbl}
\usepackage{array}
\usepackage{multicol}
\usepackage{tabularx}
\usepackage{graphicx}

\usepackage{times}
\usepackage{latexsym}

\usepackage[T1]{fontenc}

\usepackage[utf8]{inputenc}

\usepackage{microtype}

\usepackage{inconsolata}

\title{Deep Learning-Based Knowledge Injection for Metaphor Detection: A Comprehensive Review}

\author{Cheng Yang, Zheng Li, Zhiyue Liu, Qingbao Huang \\
        Guangxi University, Guangxi, China \\
        \{2212391065, 2312392066\}@st.gxu.edu.cn, liuzhy@gxu.edu.cn, qbhuang@gxu.edu.cn\\
}

\begin{document}
\maketitle
\begin{abstract}
Metaphor as an advanced cognitive modality works by extracting familiar concepts in the target domain in order to understand vague and abstract concepts in the source domain. This helps humans to quickly understand and master new domains and thus adapt to changing environments. With the continuous development of metaphor research in the natural language community, many studies using knowledge-assisted models to detect textual metaphors have emerged in recent years. Compared to not using knowledge, systems that introduce various kinds of knowledge achieve greater performance gains and reach SOTA in a recent study. Based on this, the goal of this paper is to provide a comprehensive review of research advances in the application of deep learning for knowledge injection in metaphor detection tasks. We will first systematically summarize and generalize the mainstream knowledge and knowledge injection principles. Then, the datasets, evaluation metrics, and benchmark models used in metaphor detection tasks are examined. Finally, we explore the current issues facing knowledge injection methods and provide an outlook on future research directions.
\end{abstract}

\section{Introduction}

Metaphors are essentially cognitive mechanisms present in the human mind used to construct conceptual frameworks \cite{lakoff2012little}. This phenomenon works by extracting familiar concepts in the target domain to understand vague and abstract concepts in the source domain \cite{lakoff2008metaphors}. As an important linguistic phenomenon, automatic detection of metaphors is crucial for many practical language processing tasks, including information extraction \cite{Tsvetkov_Mukomel_Gershman_2013}, sentiment analysis \cite{cambria2017sentiment}, machine translation \cite{Babieno_Takeshita_Radisavljevic_Rzepka_Araki_2022}, and seamless human-computer interaction \cite{Rai_Chakraverty_2021}. In the philosophical account articulated in \cite{294553fa-2132-3fcf-a296-3607505e8e1f}, metaphor comprehension involves three distinct phases: comprehension of the literal interpretation, discovery of inconsistencies with the literal interpretation, and reasoning to recover the intended non-literal interpretation. This paper focuses on the metaphor detection task, the first two stages of metaphor comprehension in \cite{294553fa-2132-3fcf-a296-3607505e8e1f}. Consider a metaphor detection task example: 
\begin{center}
\begin{tabular}{l}
    He's the \textbf{miracle} of the team.
\end{tabular}
\end{center}
The word "miracle" in the sentence is a metaphorical usage, which conveys something extraordinary about the person's work or collaboration by associating the word "miracle" with the person's outstanding performance in the team. In deep learning, metaphor detection is the process of determining whether a target word (e.g., "miracle") is a metaphorical usage, given the target word and context.

As the research on metaphor continues to deepen, more and more types of knowledge and injection methods have been actively explored. \citet{Mao_Lin_Guerin_2019} used general corpus information as the context of the words with detection. \citet{Le_Thai_Nguyen_2020} attempted to apply dependency tree knowledge to metaphor detection by constructing a graph network adjacency matrix to utilize the dependency tree structural information. \citet{Su_Fukumoto_Huang_Li_Wang_Chen_2020} used a prompting approach to transform metaphor detection to reading comprehension and introduced local textual information. \citet{Choi_Lee_Choi_Park_Lee_Lee_Lee_2021} took into account the basic and contextual meanings of target words. Recently, \citet{zhang2023adversarial} successfully reached the state-of-the-art of the current metaphor detection task by introducing adversarial learning and multi-task learning. The success of these studies highlights the important role of knowledge injection in systematic detection of metaphors. Compared to not using knowledge, systems that introduced various kinds of knowledge realized greater performance improvements. Therefore, we believe it is necessary to comprehensively review and summarize the current metaphor detection systems as well as the contents and approaches of knowledge injection, in order to provide theoretical foundations and model references for future researchers engaged in the exploration of new knowledge.

Although several research surveys on metaphor detection have existed in the past. \citet{Rai_Chakraverty_2021, ptivcek2023methods} provided an overview of metaphor theory and computational processing methods, \citet{Abulaish_Kamal_Zaki_2020} surveyed six technical approaches to metaphorical language, and \citet{Tong_Shutova_Lewis_2021} delved into metaphor processing methods and their applications. However, none of these surveys has taken the principle of knowledge infusion as a primary research focus. Against this background, our surveys aim to fill the gap in this research area. First, we systematically sorted out the mainstream knowledge methods and knowledge injection principles, and used an innovative categorization method to organically integrate these studies. Second, we conducted an exhaustive review and analysis of the current major metaphor datasets, including their different variants, assessment metrics, and benchmarks. Finally, we provided insights into the strengths and limitations of different knowledge injection methods, and offered suggestions and outlooks for future metaphor detection research.

\section{Knowledge}
In this section, we provide an introduction to the types of knowledge that are commonly used in metaphor detection task and how they are used. 

\subsection{Syntactic Knowledge}
\textbf{Part-of-Speech Tagging.} Part-of-Speech (POS) is the tagging of each word in a sentence to indicate its grammatical role or lexical category in the context. Commonly used POS tag sets include Universal POS tag sets \cite{Petrov_Das_McDonald_2011}, which defined a simplified set of lexical tokens with 17 tokens, such as NOUN (noun), VERB (verb), and Treebank tag sets \cite{Santorini_1990}, which had more detailed tokens, including JJ (adjective), JJS (adjective with a supreme ending -est), etc. In metaphor detection task, researchers usually combine POS knowledge directly into the input sequence \cite{Song_Zhou_Fu_Liu_Liu_2021, feng2022s}, or construct multitask learning with POS as an auxiliary task \cite{Le_Thai_Nguyen_2020}. 
\\
\textbf{Dependency Tree.} A Dependency Tree (DT) is a syntactic structural tree used to efficiently represent dependency relationships between words in a sentence. In a Dependency Tree, each word is given a node and is connected by edges to represent the directional relationship from the dependent (subordinate) word to its main dependent (head of the subordinate) word. In metaphor detection task, researchers often utilize dependency tree knowledge to improve the syntactic comprehension of their models. \citet{Le_Thai_Nguyen_2020} employed Graph Convolutional Network (GCN), which used the dependency tree knowledge as an adjacency matrix to build a graphical structure of dependency relationships between words. Some studies \cite{Song_Zhou_Fu_Liu_Liu_2021, feng2022s}, on the other hand, focused on extracting subject-verb-object relationships in dependency trees to aid in metaphor detection. \citet{Song_Zhou_Fu_Liu_Liu_2021} processed the output of subject-verb-object correspondences in text by combining, averaging, and maximizing to further capture the associations between structural semantics, while \citet{feng2022s} used a BERT Decoder \cite{Devlin_Chang_Lee_Toutanova_2019} to allow the model to generate the start and end positions of subject-predicate-objects based on the context.
\\
\subsection{Semantic Knowledge}
\textbf{VerbNet.} VerbNet \cite{schuler2005verbnet} is a verb categorization database containing nearly 4,000 English verb lemmas, and its category design refers to the study of Levin \cite{Somers_1994}. In VerbNet, each verb is attributed to one or more categories that describe the semantic roles of the verb, syntactic constraints, and semantic relations between different categories, etc. VerbNet provides two main categorization approaches: based on syntactic structure and based on predicate meaning. In the metaphor detection task, researchers \cite{Gong_Gupta_Jain_Bhat_2020, Beigman_Klebanov_Leong_Gutierrez_Shutova_Flor_2016} used VerbNet's class information to convert each lexical unit into a binary feature vector. 
\\
\textbf{FrameNet.} The main goal of FrameNet \cite{Baker_Fillmore_Lowe_1998, Lowe_1997} is to provide sentences with semantic and syntactic annotations for a large part of the vocabulary in contemporary English. The corpus of this resource is built on The British National Corpus \cite{bnc2007british}. FrameNet employs a semantic description based on frames, each of which represents a semantic concept and describes the events, participants, attributes, relations, etc. associated with that concept. The project \cite{Fillmore_Baker_Sato_2002} is an extended version of FrameNet, which adds the US National Corpus resources. In the metaphor detection, \citet{Li_Wang_Lin_Guerin_Barrault_2023} used the FrameNet provided by \cite{Fillmore_Baker_Sato_2002} in the task for frame prediction of target and contextual lexical units, and the prediction results will aid in metaphorical analysis.
\\
\textbf{WordNet.} WordNet \cite{miller1995wordnet, fellbaum1998wordnet} is a hierarchically structured lexical database in which each word forms links with other related words to represent the semantic connections between them. In the metaphor detection task, \citet{Gong_Gupta_Jain_Bhat_2020, Beigman_Klebanov_Leong_Gutierrez_Shutova_Flor_2016} classified words into fifteen categories based on the semantic links between words in WordNet and converted these categories into binary feature vectors. Such feature vectors can be used to assist the metaphor detection and improve the performance of the model. And \citet{zhang2023adversarial} considered the first of the WordNet example sentences as literal meanings and used it for multi-task learning.
\\
\textbf{Dictionary Knowledge.} Dictionary example sentences or paraphrase knowledge are intended to provide the model with knowledge of the polysemous and metaphorical meanings of the words to be detected, and help the model better understand the semantic changes and metaphorical expressions of the words to be detected in different contexts. In metaphor detection, some researchers have utilized lexical examples to extract the context-based basic meanings of the words to be detected \cite{Zhang_Liu, zhang2023adversarial}, instead of the traditional approach of directly using the words to be detected as the basic meanings. \citet{Su_Wu_Chen_2021} combined the lexical paraphrase information into the model input to achieve knowledge fusion.
\\
\textbf{Concreteness.} Concreteness is the degree to which a word is characterized by the meaning it expresses in a language. In metaphor detection, researchers often relyed on the word specificity rating dataset \cite{Brysbaert_Warriner_Kuperman_2014}. This dataset was based on the SUBTLEX-US corpus \cite{Brysbaert_New_2009} and covers 37,058 token-level samples. This dataset was rated using a 5-point scale from abstract to concrete, and the data was collected with the help of Internet crowdsourcing. In previous studies \cite{klebanov2014different, Gong_Gupta_Jain_Bhat_2020, Beigman_Klebanov_Leong_Gutierrez_Shutova_Flor_2016}, the lexical units to be detected were transformed into binary feature vectors depending on their specificity ratings.
\\
\textbf{Topic.} Using the Latent Dirichlet Allocation (LDA) model \cite{blei2003latent}, research scholars extracted a model containing 100 topics from the New York Times (NYT) corpus \cite{Sandhaus_2008} to characterize general topics discussed by the public. In the metaphor detection task, previous research work \cite{klebanov2014different, Gong_Gupta_Jain_Bhat_2020, Beigman_Klebanov_Leong_Gutierrez_Shutova_Flor_2016} matched and associated the words in each instance with these 100 topics, followed by the calculation of probability scores for each word under each topic.

\begin{table*}[h]
\centering
\small
\renewcommand{\arraystretch}{1.5}
\setlength{\tabcolsep}{1.5pt}
\begin{tabular}{ccccccc}
\toprule
\textbf{SK} & \textbf{SYK} & \textbf{EK} & \textbf{Injection Method
} & \textbf{Core Structure} & \textbf{Papers} \\

\midrule
& & & Output Modulation & $V_{S,k}=f_b(S)[k], V_k=f_b(w_k), V_S=f_b(S)[0]$ & \multirow{2}{*}{\parbox{2.8cm}{\cite{Choi_Lee_Choi_Park_Lee_Lee_Lee_2021} \newline \cite{Li_Wang_Lin_Guerin_Barrault_2023} \newline \cite{Wang_Li_Lin_Barrault_Guerin_2023} \newline \cite{Zhang_Liu}}} \\
\multirow{-2}{*}{\checkmark} & & &  \multicolumn{3}{p{11cm}}{\cellcolor{gray!10}Determine whether the target word has a literal meaning shift in context. $V_{S,k}$ and $V_k$ are the contextual and underlying meanings, respectively, and $V_s$ is the contextual.} \\

\midrule
& & & Output Modulation & $h'_i= \lambda_i\sum h_i$ & \multirow{2}{*}{\parbox{2.8cm}{\cite{Song_Zhou_Fu_Liu_Liu_2021} \newline \cite{Wang_Li_Lin_Barrault_Guerin_2023}}} \\
& \multirow{-2}{*}{\checkmark} &  & \multicolumn{3}{p{11cm}}{\cellcolor{gray!10}Assign weight $\lambda_i$ to the $i$th output token $h_i$ based on the knowledge of dependency tree.} \\

\midrule
& & & Output Modulation & $h'_t = f_d(w’_{x<t},H),\; t\in [1,7]$ & \multirow{2}{*}{\parbox{2.8cm}{\cite{feng2022s}}} \\
& \multirow{-2}{*}{\checkmark} & & \multicolumn{3}{p{11cm}}{\cellcolor{gray!10}Let the model generate the position $h'_t$ of the corresponding subject and object of the target word in the sentence, and $w'_{x<t}$ for the already predicted results.} \\

\midrule
& & & Additional Inputs & $S=[cls]s_0\oplus s_1 \oplus ... \oplus s_k[seq]$ & \multirow{2}{*}{\parbox{2.8cm}{\cite{Gong_Gupta_Jain_Bhat_2020} \newline \cite{Su_Fukumoto_Huang_Li_Wang_Chen_2020} \newline \cite{Su_Wu_Chen_2021}}} \\
\multirow{-2}{*}{\checkmark} & \multirow{-2}{*}{\checkmark} & & \multicolumn{3}{p{11cm}}{\cellcolor{gray!10}Combined text and knowledge input. $\oplus$ is the concat operation, $s_i (0\le i\le k)$ is the context or knowledge.} \\

\midrule
& & & Additional Inputs & $S=[cls]s_0[seq],\quad S' = [cls]s_1[seq]$ & \multirow{2}{*}{\parbox{2.8cm}{\cite{Zhang_Liu} \newline \cite{Babieno_Takeshita_Radisavljevic_Rzepka_Araki_2022}\newline \cite{Li_Wang_Lin_Frank_2023}}} \\
\multirow{-2}{*}{\checkmark} & & &  \multicolumn{2}{p{11cm}}{\cellcolor{gray!10}Text and knowledge will be input to different Encoders. where $s_0$ is the input text and $s_1$ is the example sentence or word paraphrase.}\\

\midrule
& & & Multi-task learning & $H,Z=G(f_b(s),f_b(s'))$ & \multirow{3}{*}{\parbox{2.8cm}{\cite{Le_Thai_Nguyen_2020} \newline \cite{Mao_Li_2021} \newline \cite{Mao_Li_Ge_Cambria_2022} \newline \cite{Li_Wang_Lin_Guerin_Barrault_2023} \newline \cite{zhang2023adversarial}}} \\
\multirow{-2}{*}{\checkmark} & & \multirow{-2}{*}{\checkmark} &  \multicolumn{3}{p{11cm}}{\cellcolor{gray!10}Multiple task samples are used and models share parameters. $H, Z$ correspond to the main task and subtask sample outputs, respectively, and G(x) is the fully connected layer, GCN, or gated network.} \\

\midrule
& & & Multi-task learning & $H^j_0=f_b(g_k^j)[0]$ & \multirow{2}{*}{\parbox{2.8cm}{\cite{Wan_Lin_Du_Shen_Zhang_2021}}} \\
\multirow{-2}{*}{\checkmark} & & &  \multicolumn{3}{p{11cm}}{\cellcolor{gray!10}A WSD subtask is designed to predict the correct meaning of the target word on the input text. $g_k^j$ is the $j$th meaning of target word $w_k$} \\

\bottomrule
\end{tabular}
\caption{Abstract of metaphor detection system based on knowledge injection. SK: semantic knowledge. SYK: syntactic knowledge. EK: emotional knowledge. core structure: subject model architecture.}
\label{tab: systems}
\end{table*}

\subsection{Emotional Knowledge}
\textbf{VAD Model.} VAD \cite{Mehrabian_1996} is an affective classification system for describing and measuring the three main dimensions of human affective experience: valence, arousal, and dominance. EmoBank corpus \cite{Buechel_Hahn_2017} is a VAD model-based and balanced multi-type 10k English corpus of sentences, each labeled with one to five ratings on the three VAD dimensions. In the metaphor detection task, \citet{Dankers_Rei_Lewis_Shutova_2019} introduced the EmoBank corpus \cite{Buechel_Hahn_2017} as an auxiliary task. Sentence-level sentiment regression was constructed based on each dimension in EmoBank. In its training process, a batch of metaphor or sentiment task data sampling is randomly selected for training at each step.
\\
\textbf{Hyperbole Corpus.} Exaggeration usually involves over- or under-exaggeration of an emotion, sentiment or attitude. Combining a dataset for hyperbole detection with a metaphor detection task can make the model more sensitive to capturing emotions and sentiments in text. In a previous research, \citet{Badathala_Kalarani_Siledar_Bhattacharyya_2023} introduced two hyperbole corpora, named HYPO and HYPO-L, and subsequently labeled them with metaphors. The results showed that multitask learning based on hyperbole and metaphor gains in both two-way performance.

\section{Method}
\label{sec:model}
This section will comprehensively introduce the current mainstream knowledge injection methods. Table \ref{tab: systems} demonstrates a summary of knowledge injection-based metaphor detection systems.

\subsection{Output Modulation}
Pre-defined knowledge can structure the tuning of model output to direct its attention to specific semantic content or syntactic structures. In many studies \cite{Mao_Lin_Guerin_2019, Choi_Lee_Choi_Park_Lee_Lee_Lee_2021, Li_Wang_Lin_Guerin_Barrault_2023, Wang_Li_Lin_Barrault_Guerin_2023, Zhang_Liu}, the main output modulation methods used are Metaphor Identification Program (MIP) and Selection Preference Violation (SPV).

MIP (Metaphor Identification Program) was originally introduced by \citet{Group_2007}. Its core logic consists in comparing the difference between a lexical unit in its original meaning and its meaning in context. For the text input $S=([cls]w_0,... ,w_k,... .w_n[seq])$ and the test word $w_k$, the contextual and underlying meanings of the lexical units are defined, respectively:
\begin{equation}
    \begin{aligned}
        & V_{S,k}= f_b(S)[k] \\
        & V_k = f_b(w_k),
    \end{aligned}
\end{equation}
where $f_b$ denotes the Encoder and $V_{S,k}$ is the corresponding output of the kth hidden layer in the text to be detected.

SPV (Selectional Preference Violation) was originally introduced to the field of metaphor detection by \citet{Wilks_Dalton_Allen_Galescu_2013}, and its core logic lies in comparing the semantic differences between the contextual meaning of a word and its surrounding words. The semantic information of surrounding words can be defined as:
\begin{align}
    V_S=f_b(S)[0],
\end{align}
where $f_b$ denotes the Encoder and $V_S$ denotes the corresponding hidden layer output of cls in the text to be detected. And the semantic information of SPV lexical units is similar to MIP as $V_{S,t}$.

In addition, \citet{Wang_Li_Lin_Barrault_Guerin_2023, Song_Zhou_Fu_Liu_Liu_2021, feng2022s} apply dependency tree knowledge to adjust the outputs. In contrast, \cite{Wang_Li_Lin_Barrault_Guerin_2023, Song_Zhou_Fu_Liu_Liu_2021} assigns different weights to the model's outputs based on the dependency tree. For the output feature $H=(h_1, h_2, \ldots, h_n)$:
\begin{align}
    h'_i= \lambda_i\sum h_i,
\end{align}
where $\lambda_i$ is the corresponding weight of the $i$th token hidden layer output. In \cite{Wang_Li_Lin_Barrault_Guerin_2023}, the weight $\lambda_i$ denotes the reciprocal of the distance between the $i$th context word and the target word. In contrast, \citet{Song_Zhou_Fu_Liu_Liu_2021} only focuses on the subject, object and underlying meaning of the target word, i.e., the subject and object of the target verb and the target word itself correspond to a weight of 1, while the rest of the contexts have a weight of 0. 

\citet{feng2022s} also considered the subject and object of the target word, but allowed the model to predict the position of the subject-predicate-object in the sentence:
\begin{equation}
    h'_t = f_d(w'_{x<t},H),\; t\in [1,7], 
\end{equation}
where $w'_{x<t}$ is the already predicted result. $h'_t$ is the $t$th predicted output, $t\in[1,6]$ corresponds to the indexes of the beginning and the end of the subject-predicate-object in sentence $H$, respectively, and $h'_7$ is the result of metaphorical classification.

\subsection{Additional Inputs}
This type of approach aims to enhance the model's understanding of the context by feeding knowledge into the model along with the text to be detected. In the metaphor detection task, the researcher introduces example sentences \cite{Zhang_Liu, Li_Wang_Lin_Frank_2023}, word paraphrases \cite{Su_Wu_Chen_2021, Babieno_Takeshita_Radisavljevic_Rzepka_Araki_2022}, or other knowledge \cite{Gong_Gupta_Jain_Bhat_2020, Su_Fukumoto_Huang_Li_Wang_Chen_2020} as additional inputs of knowledge. For input S, \citet{Su_Wu_Chen_2021, Gong_Gupta_Jain_Bhat_2020, Su_Fukumoto_Huang_Li_Wang_Chen_2020} combines the knowledge to the input:
\begin{align}
    S=[cls]s_0\oplus s_1 \oplus ... \oplus s_k[seq],
\end{align}
where $s_0$ is the input text, $\oplus$ is the concat operation, and $s_i (0<i\le k)$ is $k$ paraphrases or other knowledge. And \citet{Zhang_Liu, Li_Wang_Lin_Frank_2023, Babieno_Takeshita_Radisavljevic_Rzepka_Araki_2022} inputs the sentence to be detected separately from the knowledge:
\begin{align}
    S=[cls]s_0[seq],\quad S' = [cls]s_1[seq],
\end{align}
where $s_0,s_1$ are the input text and knowledge, respectively.

\subsection{Multi-task Learning}
Introducing other associated tasks can effectively promote knowledge fusion between tasks, thus helping to improve metaphor detection performance. For any metaphor and subtask sample input $s=(w_0,... ,w_n), s'=(w'_0,... ,w'_n)$, \citet{Le_Thai_Nguyen_2020, zhang2023adversarial, Mao_Li_2021, Mao_Li_Ge_Cambria_2022, Badathala_Kalarani_Siledar_Bhattacharyya_2023, Li_Wang_Lin_Guerin_Barrault_2023} attempts to fuse different task knowledge, which has:
\begin{align}
    H,Z=G(f_b(s),f_b(s')),
\end{align}
where $G(x)$ denotes the layer transformation function, which can be a fully connected layer, a GCN, or a gated network. $f_b$ is the Encoder, and $H$ and $Z$ denote the corresponding hidden layer outputs of the metaphorical task and other subtask samples, respectively. They are distinguished by the fact that similarity is used in \cite{Le_Thai_Nguyen_2020} for information alignment, with the loss defined as:
\begin{equation}
    \mathcal{L}=\lambda||H-Z||,
\end{equation}
the rest label the results directly:
\begin{equation}
    \begin{aligned}
        \mathcal{L}'_1=L_{CE}(\hat{y}_1,d_1) \quad \hat{y}_1\in H\\
        \mathcal{L}'_2=L_{CE}(\hat{y}_2,d_2) \quad \hat{y}_2\in Z,\\
    \end{aligned}
\end{equation}
where $d_1$ and $d_2$ are the true labels of the two tasks, respectively. $L'_1$ is a metaphor labeling task, while the definition of $L'_2$ subtasks varies across studies. $L'_2$ can be defined as a WSD task \cite{zhang2023adversarial}, a POS labeling task \cite{Mao_Li_2021, Mao_Li_Ge_Cambria_2022}, an emotion labeling task \cite{Dankers_Rei_Lewis_Shutova_2019}, an exaggeration labeling task \cite{Badathala_Kalarani_Siledar_Bhattacharyya_2023} or a FrameNet labeling task \cite{Li_Wang_Lin_Guerin_Barrault_2023}.

\begin{table*}[htbp]
\centering
\small
\setlength{\tabcolsep}{2pt}
\renewcommand{\arraystretch}{1.2}
\begin{tabular}{@{}lllllllllllllllll@{}}
\hline
& \multicolumn{4}{c}{VUA ALL} & \multicolumn{4}{c}{VUA Verb} & \multicolumn{4}{c}{MOH-X (10 fold)} & \multicolumn{4}{c}{TroFi (10 fold)} \\
\cline{2-17}
& Pre.  & Rec.  & F1   & Acc. & Pre.  & Rec.  & F1    & Acc. & Pre.    & Rec.   & F1     & Acc.   & Pre.    & Rec.   & F1     & Acc.   \\
\hline
\cite{Gao_Choi_Choi_Zettlemoyer_2018}       &       &       &      &      & 53.4  & 65.6  & 58.9  & 69.1 & 75.3    & 84.3   & 79.1   & 78.5   & 68.7    & 74.6   & 72     & 73.7   \\
\cite{Gao_Choi_Choi_Zettlemoyer_2018}  & 71.6  & 73.6  & 72.6 & 93.1 & 68.2  & 71.3  & 69.7  & 81.4 & 79.1    & 73.5   & 75.6   & 77.2   & 70.1    & 71.6   & 71.1   & 74.6   \\
\cite{Gao_Choi_Choi_Zettlemoyer_2018}  & 71.5  & 71.9  & 71.7 & 92.9 & 66.7  & 71.5  & 69    & 80.7 & 75.1    & 81.8   & 78.2   & 78.1   & 70.3    & 67.1   & 68.7   & 73.4   \\
\cite{Mao_Lin_Guerin_2019}      & 71.8  & 76.3  & 74   & 93.6 & 69.3  & 72.3  & 70.8  & 82.1 & 79.7    & 79.8   & 79.8   & 79.7   & 67.4    & 77.8   & 72.2   & 74.9   \\
\cite{Mao_Lin_Guerin_2019}   & 73    & 75.7  & 74.3 & 93.8 & 66.3  & 75.2  & 70.5  & 81.8 & 77.5    & 83.1   & 80     & 79.8   & 68.6    & 76.8   & 72.4   & 75.2   \\
\cite{Gong_Gupta_Jain_Bhat_2020}        & 74.6  & 71.5  & 73   &      & 76.7  & 77.2  & 77    &      &         &        &        &        & 72.6    & 67.5   & 69     & \\
\cite{Le_Thai_Nguyen_2020}     & 74.8  & 75.5  & 75.1 &      & 72.5  & 70.9  & 71.7  & 83.2 & 79.7    & 80.5   & 79.6   & 79.9   & 73.1    & 73.6   & 73.2   & 76.4   \\
\cite{Rohanian_Rei_Taslimipoor_Ha_2020}  &       &       &      &      &       &       &       &      & 80      & 80.4   & 80.2   & 80.5   & 73.8    & 71.8   & 72.8   & 73.5   \\
\cite{Leong_Beigman_Klebanov_Hamill_Stemle_Ubale_Chen_2020}    & 80.4  & 74.9  & 77.5 &      & 79.2  & 69.8  & 74.2  &      &         &        &        &        &         &        &        &        \\
\cite{Su_Fukumoto_Huang_Li_Wang_Chen_2020}        & 82    & 71.3  & 76.3 &      & 79.5  & 70.8  & 74.9  &      & 79.9$^\dagger$  & 76.5$^\dagger$   & 77.9$^\dagger$   &       & 53.7$^\dagger$   & 72.9$^\dagger$   & 61.7$^\dagger$   &        \\
\cite{Song_Zhou_Fu_Liu_Liu_2021}         & 82.7  & 72.5  & 77.2 & 94.7 & 80.8  & 71.5  & 75.9  & 86.4 & 80      & 85.1   & 82.1   & 81.9   & 70.4    & 74.3   & 72.2   & 75.1   \\
\cite{Wan_Lin_Du_Shen_Zhang_2021}     & 82.5  & 72.5  & 77.2 & 94.7 & 78.9  & 70.9  & 74.7  & 85.4 &         &        &        &        &   &   &  &\\
\cite{Choi_Lee_Choi_Park_Lee_Lee_Lee_2021}        & 80.1  & 76.9  & 78.5 &      & 78.7  & 72.9  & 75.7  &      & 79.3$^\dagger$    & 79.7$^\dagger$   & 79.2$^\dagger$   &        & 53.4$^\dagger$    & 74.1$^\dagger$   & 62$^\dagger$     &        \\
\cite{Li_Wang_Lin_Guerin_Barrault_2023}      &  82.7  &  75.3  &  78.8    &      &       &       &       &      & 83.2$^\dagger$    & 84.4$^\dagger$   & 83.8$^\dagger$   &        & 70.7$^\dagger$    & 78.2$^\dagger$   & 74.2$^\dagger$  & \\
\cite{Babieno_Takeshita_Radisavljevic_Rzepka_Araki_2022}       &   79.3  &  78.5  &  78.9   &      & 60.9  & 77.7  & 68.3  &      & 81      & 80     & 80.2   &        & 53.2    & 72.8   & 61.4   &        \\
\cite{Lin_Ma_Yan_Chen_2021}           & 79.3  & 78.8  & 79   & 94.8 & 78.1  & 73.2  & 75.6  & 85.8 & 85.7    & 84.6   & 84.7   & 85.2   & 74.4    & 74.8   & 74.5   & 77.7   \\
\cite{Wang_Li_Lin_Barrault_Guerin_2023}        & 80  & 78.2  & 79.1  &      &       &       &       &      & 77$^*$    & 83.5$^*$   & 80.1$^*$   &        & 54.2$^*$    & 76.2$^*$   & 63.3$^*$   &   \\
\cite{Zhang_Liu}        & 80.4  & 78.4  & 79.4 & 94.9 & 78.3  & 73.6  & 75.9  & 86   & 84      & 84     & 83.4   & 83.6   & 67.5    & 77.6   & 71.9   & 73.6   \\
\cite{feng2022s}         & 81.6  & 77.4  & 79.4 & 95.2 & 81.6  & 71.1  & 76    & 86.4 & 89.5    & 85.2   & 87     & 87.5   & 72.5    & 77.5   & 74.8   & 77.7   \\
\cite{Su_Wu_Chen_2021} &       &       &      &      & 76    & 76    & 76    & 85.7 & 82.9    & 84     & 83.4   & 84.2   & 73.3    & 69.6   & 71.4   & 75.7   \\
\cite{zhang2023adversarial}          & 78.4  & 79.5  & 79   & 94.7 & 78.5  & 78.1  & 78.3  & 87   & 87.4    & 88.8   & 87.9   & 88     & 70.5    & 79.8   & 74.7   & 76.5   \\
\hline
\end{tabular}
\caption{This table shows the performance of the metaphor detection system on four datasets, VUA ALL, VUA verb, MOH-X and TroFi, in recent years. Among them, most of the results on the MOH-X and TroFi datasets are based on ten-fold cross-validation, and also include some results derived from direct computation ($^\dagger$ labeling), as well as some of the models are trained on the VUA20 dataset ($^*$ labeling).
}
\label{tab: performance}
\end{table*}

\cite{Wan_Lin_Du_Shen_Zhang_2021} fine-tunes another Encoder for any metaphorical sample $s=(w_0,... ,w_k,... ,w_n)$ and the target word $w_k$, $g_k^j$ is the $j$th meaning of $w_k$, with:
\begin{align}
    H^j_0=f_b(g_k^j)[0],
\end{align}
where $H^j_0$ is the $j$th meaning of the target word $w_k$ corresponding to the CLS hidden layer output. If $g^j_k$ is the contextual meaning of the target word $w_k$, it is labeled 1, and vice versa 0.

\section{Metrics and Dataset}
The purpose of this section is to provide an overview of the currently dominant metaphor detection datasets, about which detailed information has been presented in Table \ref{tab: datasets}. The datasets will be presented in the following section. At the same time, we will introduce commonly used evaluation metrics in the field. In addition, we will summarize the performance of metaphor detection tasks performed on four datasets, namely, VUA ALL, VUA verb, MOH-X, and TroFi, in recent years, and refer to Table \ref{tab: performance} for more details, in order to provide a comprehensive picture of the state of the art of research in this area.

\subsection{Metrics}
Current mainstream metaphor detection systems typically use four evaluation metrics. Among them, accuracy indicates the number of correctly categorized samples as a proportion of the total number of samples; precision measures the extent to which the model correctly predicts, focusing on the proportion of samples that the model determines to be in the positive category that are truly in the positive category; and recall measures the model's ability to correctly identify positively categorized samples (true instances). The F1-score is a metric that combines precision and recall and is used to balance the model's accuracy and recall.

\subsection{Dataset}
\textbf{VUA.}\; The VU Amsterdam Metaphor Corpus \cite{Steen_Dorst_Herrmann_Kaal_Krennmayr_Pasma_2010} annotates each lexical unit (187,570 in total) in a subset of the British National Corpus \cite{bnc2007british} metaphorically. The corpus was tagged using the MIPVU metaphor detection program, and VUAMC is the largest publicly available annotated corpus of tag-level metaphor detection, and the only one to study the metaphorical nature of dummy words.
\\
\textbf{VUA SEQ.}\; VUA SEQ is another dataset constructed based on VUAMC. Compared to VUA ALL, VUA SEQ has the same number of samples as reported \cite{Gao_Choi_Choi_Zettlemoyer_2018, Neidlein_Wiesenbach_Markert_2020}. However, VUA SEQ covers all tokens in a sentence, even punctuation, in the classification task, thus leading to a richer number of target tokens used than VUA ALL.
\begin{table}[h]
\centering
\resizebox{0.48\textwidth}{!}{
\small
\begin{tabular}{lcccc}
\toprule
Dataset           & \#Tok. & \#Sent. & \%Met.   \\
\midrule
VUAall/SEQ     & 205,425 & 10,567 & 11.6\% \\
VUAall/SEQ/tr  & 116,622 & 6,323 & 11.2\% \\
VUAall/SEQ/val & 38,628  & 1,550 & 11.6\% \\
VUAall/SEQ/te  & 50,175  & 2,694 & 12.4\% \\
VUAallpos      & 94,807  & 16,202 & 15.8\% \\
VUAallpos\_tr  & 72,611  & 12,122 & 15.2\% \\
VUAallpos\_te  & 22,196  & 4,080 & 17.9\% \\
VUAverb\_tr    & 15,516  & 7,479 & 27.9\% \\
VUAverb\_val   & 1,724   & 1,541 & 26.9\% \\
VUAverb\_te    & 5,873   & 2,694 & 29.9\% \\
MOH-X          & 647     & 647   & 48.7\% \\
TroFi          & 3,737   & 3,737 & 43.5\% \\
\bottomrule
\end{tabular}
}
\caption{tr: training set. val: validation set. te: test set. tokens: number of samples. sent.: total number of sentences, \%Met.: percentage of metaphorical samples}
\label{tab: datasets}
\end{table}
\\
\textbf{VUA ALL POS.}\; VUA ALL POS dataset has been applied to the shared task of metaphor detection \cite{Leong_Beigman_Klebanov_Shutova_2018, Leong_Beigman_Klebanov_Hamill_Stemle_Ubale_Chen_2020}, which consists of two parts, VUA ALL POS and VUA Verb. In particular, VUA ALL POS annotates all real-sense words (including adjectives, verbs, nouns, and adjectives) in a sentence; while VUA Verb covers only verbs. However, in the studies of \cite{Song_Zhou_Fu_Liu_Liu_2021, feng2022s, Wan_Lin_Du_Shen_Zhang_2021, Su_Fukumoto_Huang_Li_Wang_Chen_2020}, the VUA ALL POS dataset also includes dummy words. To distinguish it from the shared task \cite{Leong_Beigman_Klebanov_Shutova_2018, Leong_Beigman_Klebanov_Hamill_Stemle_Ubale_Chen_2020}, we named the VUA ALL POS dataset that includes both real and dummy words as VUA ALL.
\\
\textbf{VUA Verb.}\; VUA Verb is a verb part extracted from VUA \cite{Steen_Dorst_Herrmann_Kaal_Krennmayr_Pasma_2010}. The number of training, validation, and test sets for VUA Verb are 15,516, 1,639, and 5,873, respectively, as reported in the metaphor detection shared task \cite{Leong_Beigman_Klebanov_Shutova_2018, Leong_Beigman_Klebanov_Hamill_Stemle_Ubale_Chen_2020}.
\\
\textbf{VUA18.}\; According to \cite{Choi_Lee_Choi_Park_Lee_Lee_Lee_2021}, VUA-18 is very similar to VUA-SEQ and VUA ALL as they use the same sentences in each subset, 6,323, 1,550, and 2,694 sentences for the training, development, and test sets, respectively. VUA-18 does not consider abbreviations and punctuation as separate tokens, and has the same labeling rules as VUA ALL same as VUA ALL's labeling rules, so we group VUA-18 with VUA ALL.
\\
\textbf{VUA20.}\; In the literature \cite{Choi_Lee_Choi_Park_Lee_Lee_Lee_2021}, VUA20 labeled 1.2k sentences with real and imaginary words. However, this does not match the description in the 20-year shared task \cite{Leong_Beigman_Klebanov_Hamill_Stemle_Ubale_Chen_2020}. The text states that it uses the same VUA as the 18-year shared task \cite{Leong_Beigman_Klebanov_Shutova_2018} (see Section 3.1, lines 8-10) and that both report the same token count. Given this, we will not list VUA-18 and VUA-20 in the model performance table.
\\
\textbf{TroFi.}\; TroFi \cite{Birke_Sarkar_2005} is a dataset focused on detecting verb metaphors, which contains the literal and metaphorical usage of 50 English verbs from the 1987-1989 Wall Street Journal corpus \cite{charniak2000bllip}. The dataset contains a total of 3717 samples and is not segmented.
\\
\textbf{MOH.}\; MOH \cite{Mohammad_Shutova_Turney_2016}, which also focuses on verb metaphors, consists of 1639 sentences extracted from WordNet, containing 1230 sentences with literal usage and 409 metaphorical usages, which were metaphorically labeled through crowdsourcing. MOH-X \cite{Shutova_Kiela_Maillard_2016}, on the other hand, is a subset of the MOH dataset, excluding instances with pronouns and subordinate subjects or objects, and contains 647 verb samples.

\section{Future Direction}
\subsection{Refining the Criteria for Defining Metaphors}
When introducing external knowledge, researchers often have vague criteria for defining metaphors. For example, some studies consider the first paraphrase in WordNet as the basic meaning \cite{zhang2023adversarial}, or use the first k example sentences in the dictionary as the criterion for classifying non-metaphorical expressions \cite{Su_Wu_Chen_2021, Zhang_Liu}. However, these approaches may introduce knowledge noise that negatively affects the performance of the model.
\\
\textbf{Precise Knowledge Annotation Methods.}\; Future research could be conducted through manual annotation, detailed paraphrasing in specialized field dictionaries, or the use of professional vetting to ensure the accuracy and relevance of external knowledge.
\\
\textbf{Applications of Large Pre-trained Models.}\; Utilizing large language models (e.g., GPT-3 \cite{brown2020language}) to mine the implicit knowledge learned by the model itself. This approach can provide more accurate external knowledge for metaphor detection by analyzing the model-generated text and extracting the metaphorical information in it. For example, \cite{wachowiak2023does} guided GPT-3 to generate the target domain of metaphors.
\\
\textbf{Real-Time Knowledge Update Mechanism:} Most of the past studies use older knowledge bases \cite{miller1995wordnet, schuler2005verbnet, Baker_Fillmore_Lowe_1998}, and it is necessary to consider designing a real-time knowledge update mechanism. By updating the external knowledge base on a regular or real-time basis to reflect changes in language usage and context, the model is better adapted to evolving contexts.

\subsection{Enhancing the Knowledge Infusion Methodology}
Current research has used two main approaches to inject knowledge into models: incorporating knowledge directly into the input of the model \cite{Li_Wang_Lin_Frank_2023, Babieno_Takeshita_Radisavljevic_Rzepka_Araki_2022} or adapting the output of the model \cite{Wang_Li_Lin_Barrault_Guerin_2023, feng2022s}. However, these traditional combination approaches may not fully utilize the rich contextual information embedded in the knowledge.
\\
\textbf{Exploring More Effective Ways of Knowledge Fusion.}\; Future research can further explore more effective ways of knowledge fusion to overcome the problem that traditional combining approaches cannot fully utilize contextual information. For example, \citet{Li_Wang_Lin_Guerin_Barrault_2023} introduced implicit knowledge in FrameNet by fine-tuning the model. \citet{Mao_Li_2021, Mao_Li_Ge_Cambria_2022} used a gating mechanism for knowledge fusion.
\\
\textbf{Deepening Applications of Adversarial Learning.}\; Adversarial learning \citet{zhang2023adversarial} has shown potential in knowledge fusion. Future research could deeply explore the use of adversarial learning to fuse other associative subtask knowledge, such as fine-grained sentiment information.
\\
\textbf{Combining Multiple Sources of Knowledge for Injection.}\; Current research has focused on single domain knowledge injection \cite{Badathala_Kalarani_Siledar_Bhattacharyya_2023, Dankers_Rei_Lewis_Shutova_2019}, and future research could consider how to better combine multiple sources of knowledge. For example, combining hyperbole and emotion, or emotion with other rhetorical knowledge, to create more comprehensive models that are better adapted to diverse textual expressions.

\subsection{Exploring Zero-shot Metaphor Detection}
Most metaphorical systems are trained using manually labeled data. However, data labeling requires significant labor costs and the quality of the data greatly depends on the education level of the labeler. Zero-shot metaphor detection can alleviate the above problems to some extent.
\\
\textbf{Improvement of Existing Zero-shot Methods.}\; In the past research, \cite{Mao_Lin_Guerin_2018} introduced the cosine similarity of words, and judged words larger than a certain threshold as metaphors. And \cite{Mao_Li_Ge_Cambria_2022} defined words with the highest probability of occurrence in the BERT context as non-metaphors. The above methods have certain shortcomings in terms of detection scope and accuracy, and need to be further improved so that they can detect metaphorical expressions more accurately. For example, word-level similarity measures (e.g., cosine similarity) are extended to sentence-level or document-level similarity (e.g., WMD \cite{kusner2015word}) to better capture the context of metaphors.
\\
\textbf{Generating Metaphor Datasets Using Large Models.}\; Data generation with large models has been shown to be cost-effective and efficient \cite{wang2021want, yoo2021gpt3mix}. Therefore, using large models to generate textual datasets containing metaphors helps to scale up the training data and reduce the reliance on manual annotation, while increasing the diversity and coverage of the data.
\\
\textbf{Guiding the Big Model for Metaphor Detection Via Prompt Learning.}\; Prompt learning \cite{ye2022zerogen, meng2022generating} aims to guide the LLM to generate specific content in a non-fine-tuned manner. In this task, the LLM plays the role of a few or zero sample learner. How to guide the model to deeply understand the features of metaphors and design prompts that can evoke metaphorical expressions is a very worthwhile direction for future research.

\section{Conclusion}
With the in-depth study of metaphor detection tasks, most models improve the detection of metaphors by injecting different knowledge. The role of knowledge injection in metaphor detection is becoming increasingly prominent. In this paper, we comprehensively review the knowledge contents and injection methods used by deep learning models on the metaphor detection task. We categorize the knowledge content into semantic, syntactic and affective knowledge. Meanwhile, we classify the knowledge injection methods as output moderation, additional input, and multi-task learning. Next, we introduce the commonly used datasets and evaluation metrics for metaphor detection tasks, and show the performance of the current metaphor detection system on four classical datasets, VUA ALL, VUAverb, TroFi and MOH-X. Finally, we discuss the problems of current metaphor detection systems and provide directions for future research.

\section{Limitations}
This paper provides a comprehensive description of metaphor detection systems in deep learning, focusing on discussing and summarizing in detail the different types and methods of model knowledge injection. However, there exists a small amount of research work in the area of metaphor detection that does not use knowledge or employs unsupervised methods, and these studies are not covered or discussed in the paper. In future research, we plan to provide a comprehensive summary of most of the work in the area of metaphor detection, including both supervised and unsupervised approaches, to provide researchers with a more comprehensive understanding.

\section{Ethics Statement}
In this paper, we provide a detailed description of the supervised metaphor detection system and the different ways of knowledge injection. The datasets and research papers we have used have been obtained from publicly available sources and we have adhered to strict guidelines of academic and research ethics. In addition, we place special emphasis on transparency and openness of information, encourage other researchers to conduct responsible research, and uphold best practices in knowledge sharing. In the text, we explicitly cite the public data sources cited to express our full respect for the original authors and data providers of research related to the field of metaphor detection.

\bibliography{anthology,custom}

\begin{thebibliography}{68}
\expandafter\ifx\csname natexlab\endcsname\relax\def\natexlab#1{#1}\fi

\bibitem[{Abulaish et~al.(2020)Abulaish, Kamal, and
  Zaki}]{Abulaish_Kamal_Zaki_2020}
Muhammad Abulaish, Ashraf Kamal, and Mohammed~J. Zaki. 2020.
\newblock \href {https://doi.org/10.1145/3375547} {A survey of figurative
  language and its computational detection in online social networks}.
\newblock \emph{ACM Transactions on the Web}, 14(1):1–52.

\bibitem[{Babieno et~al.(2022)Babieno, Takeshita, Radisavljevic, Rzepka, and
  Araki}]{Babieno_Takeshita_Radisavljevic_Rzepka_Araki_2022}
Mateusz Babieno, Masashi Takeshita, Dusan Radisavljevic, Rafal Rzepka, and
  Kenji Araki. 2022.
\newblock \href {https://doi.org/10.3390/app12042081} {Miss roberta wilde:
  Metaphor identification using masked language model with wiktionary lexical
  definitions}.
\newblock \emph{Applied Sciences}, 12(4):2081.

\bibitem[{Badathala et~al.(2023)Badathala, Kalarani, Siledar, and
  Bhattacharyya}]{Badathala_Kalarani_Siledar_Bhattacharyya_2023}
Naveen Badathala, AbisekRajakumar Kalarani, Tejpalsingh Siledar, and Pushpak
  Bhattacharyya. 2023.
\newblock A match made in heaven: A multi-task framework for hyperbole and
  metaphor detection.

\bibitem[{Baker et~al.(1998)Baker, Fillmore, and
  Lowe}]{Baker_Fillmore_Lowe_1998}
Collin~F. Baker, Charles~J. Fillmore, and John~B. Lowe. 1998.
\newblock \href {https://doi.org/10.3115/980845.980860} {The berkeley framenet
  project}.
\newblock In \emph{Proceedings of the 36th annual meeting on Association for
  Computational Linguistics -}.

\bibitem[{Beigman~Klebanov et~al.(2016)Beigman~Klebanov, Leong, Gutierrez,
  Shutova, and Flor}]{Beigman_Klebanov_Leong_Gutierrez_Shutova_Flor_2016}
Beata Beigman~Klebanov, Chee~Wee Leong, E.~Dario Gutierrez, Ekaterina Shutova,
  and Michael Flor. 2016.
\newblock \href {https://doi.org/10.18653/v1/p16-2017} {Semantic
  classifications for detection of verb metaphors}.
\newblock In \emph{Proceedings of the 54th Annual Meeting of the Association
  for Computational Linguistics (Volume 2: Short Papers)}.

\bibitem[{Birke and Sarkar(2005)}]{Birke_Sarkar_2005}
Julia Birke and Anoop Sarkar. 2005.
\newblock A clustering approach for nearly unsupervised recognition of
  nonliteral language.
\newblock \emph{Conference of the European Chapter of the Association for
  Computational Linguistics,Conference of the European Chapter of the
  Association for Computational Linguistics}.

\bibitem[{Blei et~al.(2003)Blei, Ng, and Jordan}]{blei2003latent}
David~M Blei, Andrew~Y Ng, and Michael~I Jordan. 2003.
\newblock Latent dirichlet allocation.
\newblock \emph{Journal of machine Learning research}, 3(Jan):993--1022.

\bibitem[{Brown et~al.(2020)Brown, Mann, Ryder, Subbiah, Kaplan, Dhariwal,
  Neelakantan, Shyam, Sastry, Askell et~al.}]{brown2020language}
Tom Brown, Benjamin Mann, Nick Ryder, Melanie Subbiah, Jared~D Kaplan, Prafulla
  Dhariwal, Arvind Neelakantan, Pranav Shyam, Girish Sastry, Amanda Askell,
  et~al. 2020.
\newblock Language models are few-shot learners.
\newblock \emph{Advances in neural information processing systems},
  33:1877--1901.

\bibitem[{Brysbaert and New(2009)}]{Brysbaert_New_2009}
Marc Brysbaert and Boris New. 2009.
\newblock \href {https://doi.org/10.3758/brm.41.4.977} {Moving beyond kučera
  and francis: A critical evaluation of current word frequency norms and the
  introduction of a new and improved word frequency measure for american
  english}.
\newblock \emph{Behavior Research Methods}, page 977–990.

\bibitem[{Brysbaert et~al.(2014)Brysbaert, Warriner, and
  Kuperman}]{Brysbaert_Warriner_Kuperman_2014}
Marc Brysbaert, Amy~Beth Warriner, and Victor Kuperman. 2014.
\newblock \href {https://doi.org/10.3758/s13428-013-0403-5} {Concreteness
  ratings for 40 thousand generally known english word lemmas}.
\newblock \emph{Behavior Research Methods}, 46(3):904–911.

\bibitem[{Buechel and Hahn(2017)}]{Buechel_Hahn_2017}
Sven Buechel and Udo Hahn. 2017.
\newblock \href {https://doi.org/10.18653/v1/e17-2092} {Emobank: Studying the
  impact of annotation perspective and representation format on dimensional
  emotion analysis}.
\newblock In \emph{Proceedings of the 15th Conference of the European Chapter
  of the Association for Computational Linguistics: Volume 2, Short Papers}.

\bibitem[{Cambria et~al.(2017)Cambria, Poria, Gelbukh, and
  Thelwall}]{cambria2017sentiment}
Erik Cambria, Soujanya Poria, Alexander Gelbukh, and Mike Thelwall. 2017.
\newblock Sentiment analysis is a big suitcase.
\newblock \emph{IEEE Intelligent Systems}, 32(6):74--80.

\bibitem[{Charniak et~al.(2000)Charniak, Blaheta, Ge, Hall, Hale, and
  Johnson}]{charniak2000bllip}
Eugene Charniak, Don Blaheta, Niyu Ge, Keith Hall, John Hale, and Mark Johnson.
  2000.
\newblock Bllip 1987-89 wsj corpus release 1.
\newblock \emph{Linguistic Data Consortium, Philadelphia}, 36.

\bibitem[{Choi et~al.(2021)Choi, Lee, Choi, Park, Lee, Lee, and
  Lee}]{Choi_Lee_Choi_Park_Lee_Lee_Lee_2021}
Minjin Choi, Sunkyung Lee, Eunseong Choi, Heesoo Park, Junhyuk Lee, Dongwon
  Lee, and Jongwuk Lee. 2021.
\newblock \href {https://doi.org/10.18653/v1/2021.naacl-main.141} {Melbert:
  Metaphor detection via contextualized late interaction using metaphorical
  identification theories}.
\newblock In \emph{Proceedings of the 2021 Conference of the North American
  Chapter of the Association for Computational Linguistics: Human Language
  Technologies}.

\bibitem[{Consortium et~al.(2007)}]{bnc2007british}
BNC Consortium et~al. 2007.
\newblock British national corpus.
\newblock \emph{Oxford Text Archive Core Collection}.

\bibitem[{Dankers et~al.(2019)Dankers, Rei, Lewis, and
  Shutova}]{Dankers_Rei_Lewis_Shutova_2019}
Verna Dankers, Marek Rei, Martha Lewis, and Ekaterina Shutova. 2019.
\newblock \href {https://doi.org/10.18653/v1/d19-1227} {Modelling the interplay
  of metaphor and emotion through multitask learning}.
\newblock In \emph{Proceedings of the 2019 Conference on Empirical Methods in
  Natural Language Processing and the 9th International Joint Conference on
  Natural Language Processing (EMNLP-IJCNLP)}.

\bibitem[{Devlin et~al.(2019)Devlin, Chang, Lee, and
  Toutanova}]{Devlin_Chang_Lee_Toutanova_2019}
Jacob Devlin, Ming-Wei Chang, Kenton Lee, and Kristina Toutanova. 2019.
\newblock \href {https://doi.org/10.18653/v1/n19-1423} {Bert: Pre-training of
  deep bidirectional transformers for language understanding}.
\newblock In \emph{Proceedings of the 2019 Conference of the North}.

\bibitem[{Fellbaum(1998)}]{fellbaum1998wordnet}
Christiane Fellbaum. 1998.
\newblock \emph{WordNet: An electronic lexical database}.
\newblock MIT press.

\bibitem[{Feng and Ma(2022)}]{feng2022s}
Huawen Feng and Qianli Ma. 2022.
\newblock It’s better to teach fishing than giving a fish: An auto-augmented
  structure-aware generative model for metaphor detection.
\newblock In \emph{Findings of the Association for Computational Linguistics:
  EMNLP 2022}, pages 656--667.

\bibitem[{Fillmore et~al.(2002)Fillmore, Baker, and
  Sato}]{Fillmore_Baker_Sato_2002}
CharlesJ. Fillmore, CollinF. Baker, and Hiroaki Sato. 2002.
\newblock The framenet database and software tools.
\newblock \emph{Language Resources and Evaluation,Language Resources and
  Evaluation}.

\bibitem[{Gao et~al.(2018)Gao, Choi, Choi, and
  Zettlemoyer}]{Gao_Choi_Choi_Zettlemoyer_2018}
Ge~Gao, Eunsol Choi, Yejin Choi, and Luke Zettlemoyer. 2018.
\newblock \href {https://doi.org/10.18653/v1/d18-1060} {Neural metaphor
  detection in context}.
\newblock In \emph{Proceedings of the 2018 Conference on Empirical Methods in
  Natural Language Processing}.

\bibitem[{Gong et~al.(2020)Gong, Gupta, Jain, and
  Bhat}]{Gong_Gupta_Jain_Bhat_2020}
Hongyu Gong, Kshitij Gupta, Akriti Jain, and Suma Bhat. 2020.
\newblock \href {https://doi.org/10.18653/v1/2020.figlang-1.21} {Illinimet:
  Illinois system for metaphor detection with contextual and linguistic
  information.}
\newblock In \emph{Proceedings of the Second Workshop on Figurative Language
  Processing}.

\bibitem[{Group(2007)}]{Group_2007}
Pragglejaz Group. 2007.
\newblock \href {https://doi.org/10.1080/10926480709336752} {Mip: A method for
  identifying metaphorically used words in discourse}.
\newblock \emph{Metaphor and Symbol}, page 1–39.

\bibitem[{Klebanov et~al.(2014)Klebanov, Leong, Heilman, and
  Flor}]{klebanov2014different}
Beata~Beigman Klebanov, Ben Leong, Michael Heilman, and Michael Flor. 2014.
\newblock Different texts, same metaphors: Unigrams and beyond.
\newblock In \emph{Proceedings of the Second Workshop on Metaphor in NLP},
  pages 11--17.

\bibitem[{Kusner et~al.(2015)Kusner, Sun, Kolkin, and
  Weinberger}]{kusner2015word}
Matt Kusner, Yu~Sun, Nicholas Kolkin, and Kilian Weinberger. 2015.
\newblock From word embeddings to document distances.
\newblock In \emph{International conference on machine learning}, pages
  957--966. PMLR.

\bibitem[{Lakoff and Johnson(2008)}]{lakoff2008metaphors}
George Lakoff and Mark Johnson. 2008.
\newblock \emph{Metaphors we live by}.
\newblock University of Chicago press.

\bibitem[{Lakoff and Wehling(2012)}]{lakoff2012little}
George Lakoff and Elisabeth Wehling. 2012.
\newblock \emph{The little blue book: The essential guide to thinking and
  talking democratic}.
\newblock Simon and Schuster.

\bibitem[{Le et~al.(2020)Le, Thai, and Nguyen}]{Le_Thai_Nguyen_2020}
Duong Le, My~Thai, and Thien Nguyen. 2020.
\newblock \href {https://doi.org/10.1609/aaai.v34i05.6326} {Multi-task learning
  for metaphor detection with graph convolutional neural networks and word
  sense disambiguation}.
\newblock \emph{Proceedings of the AAAI Conference on Artificial Intelligence},
  34(05):8139–8146.

\bibitem[{Leong et~al.(2020)Leong, Beigman~Klebanov, Hamill, Stemle, Ubale, and
  Chen}]{Leong_Beigman_Klebanov_Hamill_Stemle_Ubale_Chen_2020}
Chee Wee~(Ben) Leong, Beata Beigman~Klebanov, Chris Hamill, Egon Stemle, Rutuja
  Ubale, and Xianyang Chen. 2020.
\newblock \href {https://doi.org/10.18653/v1/2020.figlang-1.3} {A report on the
  2020 vua and toefl metaphor detection shared task}.
\newblock In \emph{Proceedings of the Second Workshop on Figurative Language
  Processing}.

\bibitem[{Leong et~al.(2018)Leong, Beigman~Klebanov, and
  Shutova}]{Leong_Beigman_Klebanov_Shutova_2018}
Chee Wee~(Ben) Leong, Beata Beigman~Klebanov, and Ekaterina Shutova. 2018.
\newblock \href {https://doi.org/10.18653/v1/w18-0907} {A report on the 2018
  vua metaphor detection shared task}.
\newblock In \emph{Proceedings of the Workshop on Figurative Language
  Processing}.

\bibitem[{Li et~al.(2023{\natexlab{a}})Li, Wang, Lin, and
  Frank}]{Li_Wang_Lin_Frank_2023}
Yucheng Li, Shun Wang, Chenghua Lin, and Guerin Frank. 2023{\natexlab{a}}.
\newblock Metaphor detection via explicit basic meanings modelling.

\bibitem[{Li et~al.(2023{\natexlab{b}})Li, Wang, Lin, Guerin, and
  Barrault}]{Li_Wang_Lin_Guerin_Barrault_2023}
Yucheng Li, Shun Wang, Chenghua Lin, Frank Guerin, and Lo\"ic Barrault.
  2023{\natexlab{b}}.
\newblock Framebert: Conceptual metaphor detection with frame embedding
  learning.

\bibitem[{Lin et~al.(2021)Lin, Ma, Yan, and Chen}]{Lin_Ma_Yan_Chen_2021}
Zhenxi Lin, Qianli Ma, Jiangyue Yan, and Jieyu Chen. 2021.
\newblock Cate: A contrastive pre-trained model for metaphor detection with
  semi-supervised learning.
\newblock \emph{Empirical Methods in Natural Language Processing,Empirical
  Methods in Natural Language Processing}.

\bibitem[{Lowe(1997)}]{Lowe_1997}
JohnB. Lowe. 1997.
\newblock A frame-semantic approach to semantic annotation.

\bibitem[{Maloney(1983)}]{294553fa-2132-3fcf-a296-3607505e8e1f}
J.~Christopher Maloney. 1983.
\newblock \href {http://www.jstor.org/stable/42968961} {A new model for
  metaphor}.
\newblock \emph{Dialectica}, 37(4):285--301.

\bibitem[{Mao and Li(2021)}]{Mao_Li_2021}
Rui Mao and Xiao Li. 2021.
\newblock Bridging towers of multi-task learning with a gating mechanism for
  aspect-based sentiment analysis and sequential metaphor identification.
\newblock \emph{Proceedings of the ... AAAI Conference on Artificial
  Intelligence,Proceedings of the ... AAAI Conference on Artificial
  Intelligence}.

\bibitem[{Mao et~al.(2022)Mao, Li, Ge, and Cambria}]{Mao_Li_Ge_Cambria_2022}
Rui Mao, Xiao Li, Mengshi Ge, and Erik Cambria. 2022.
\newblock \href {https://doi.org/10.1016/j.inffus.2022.06.002} {Metapro: A
  computational metaphor processing model for text pre-processing}.
\newblock \emph{Information Fusion}, 86–87:30–43.

\bibitem[{Mao et~al.(2018)Mao, Lin, and Guerin}]{Mao_Lin_Guerin_2018}
Rui Mao, Chenghua Lin, and Frank Guerin. 2018.
\newblock \href {https://doi.org/10.18653/v1/p18-1113} {Word embedding and
  wordnet based metaphor identification and interpretation}.
\newblock In \emph{Proceedings of the 56th Annual Meeting of the Association
  for Computational Linguistics (Volume 1: Long Papers)}.

\bibitem[{Mao et~al.(2019)Mao, Lin, and Guerin}]{Mao_Lin_Guerin_2019}
Rui Mao, Chenghua Lin, and Frank Guerin. 2019.
\newblock \href {https://doi.org/10.18653/v1/p19-1378} {End-to-end sequential
  metaphor identification inspired by linguistic theories}.
\newblock In \emph{Proceedings of the 57th Annual Meeting of the Association
  for Computational Linguistics}.

\bibitem[{Mehrabian(1996)}]{Mehrabian_1996}
Albert Mehrabian. 1996.
\newblock \href {https://doi.org/10.1007/bf02686918}
  {Pleasure-arousal-dominance: A general framework for describing and measuring
  individual differences in temperament}.
\newblock \emph{Current Psychology}, page 261–292.

\bibitem[{Meng et~al.(2022)Meng, Huang, Zhang, and Han}]{meng2022generating}
Yu~Meng, Jiaxin Huang, Yu~Zhang, and Jiawei Han. 2022.
\newblock Generating training data with language models: Towards zero-shot
  language understanding.
\newblock \emph{Advances in Neural Information Processing Systems},
  35:462--477.

\bibitem[{Miller(1995)}]{miller1995wordnet}
George~A Miller. 1995.
\newblock Wordnet: a lexical database for english.
\newblock \emph{Communications of the ACM}, 38(11):39--41.

\bibitem[{Mohammad et~al.(2016)Mohammad, Shutova, and
  Turney}]{Mohammad_Shutova_Turney_2016}
SaifM. Mohammad, Ekaterina Shutova, and PeterD. Turney. 2016.
\newblock Metaphor as a medium for emotion: An empirical study.
\newblock \emph{Joint Conference on Lexical and Computational Semantics,Joint
  Conference on Lexical and Computational Semantics}.

\bibitem[{Neidlein et~al.(2020)Neidlein, Wiesenbach, and
  Markert}]{Neidlein_Wiesenbach_Markert_2020}
Arthur Neidlein, Philip Wiesenbach, and Katja Markert. 2020.
\newblock \href {https://doi.org/10.18653/v1/2020.coling-main.332} {An analysis
  of language models for metaphor recognition}.
\newblock In \emph{Proceedings of the 28th International Conference on
  Computational Linguistics}.

\bibitem[{Petrov et~al.(2011)Petrov, Das, and
  McDonald}]{Petrov_Das_McDonald_2011}
Slav Petrov, Dipanjan Das, and Ryan McDonald. 2011.
\newblock A universal part-of-speech tagset.
\newblock \emph{Cornell University - arXiv,Cornell University - arXiv}.

\bibitem[{Pti{\v{c}}ek and Dob{\v{s}}a(2023)}]{ptivcek2023methods}
Martina Pti{\v{c}}ek and Jasminka Dob{\v{s}}a. 2023.
\newblock Methods of annotating and identifying metaphors in the field of
  natural language processing.
\newblock \emph{Future Internet}, 15(6):201.

\bibitem[{Rai and Chakraverty(2021)}]{Rai_Chakraverty_2021}
Sunny Rai and Shampa Chakraverty. 2021.
\newblock \href {https://doi.org/10.1145/3373265} {A survey on computational
  metaphor processing}.
\newblock \emph{ACM Computing Surveys}, page 1–37.

\bibitem[{Rohanian et~al.(2020)Rohanian, Rei, Taslimipoor, and
  Ha}]{Rohanian_Rei_Taslimipoor_Ha_2020}
Omid Rohanian, Marek Rei, Shiva Taslimipoor, and Le~An Ha. 2020.
\newblock \href {https://doi.org/10.18653/v1/2020.acl-main.259} {Verbal
  multiword expressions for identification of metaphor}.
\newblock In \emph{Proceedings of the 58th Annual Meeting of the Association
  for Computational Linguistics}.

\bibitem[{Sandhaus(2008)}]{Sandhaus_2008}
Evan Sandhaus. 2008.
\newblock The new york times annotated corpus.

\bibitem[{Santorini(1990)}]{Santorini_1990}
Beatrice Santorini. 1990.
\newblock Part-of-speech tagging guidelines for the penn treebank project (3rd
  revision).

\bibitem[{Schuler(2005)}]{schuler2005verbnet}
Karin~Kipper Schuler. 2005.
\newblock \emph{VerbNet: A broad-coverage, comprehensive verb lexicon}.
\newblock University of Pennsylvania.

\bibitem[{Shutova et~al.(2016)Shutova, Kiela, and
  Maillard}]{Shutova_Kiela_Maillard_2016}
Ekaterina Shutova, Douwe Kiela, and Jean Maillard. 2016.
\newblock \href {https://doi.org/10.18653/v1/n16-1020} {Black holes and white
  rabbits: Metaphor identification with visual features}.
\newblock In \emph{Proceedings of the 2016 Conference of the North American
  Chapter of the Association for Computational Linguistics: Human Language
  Technologies}.

\bibitem[{Somers(1994)}]{Somers_1994}
HaroldL. Somers. 1994.
\newblock Book reviews: English verb classes and alternations: A preliminary
  investigation.

\bibitem[{Song et~al.(2021)Song, Zhou, Fu, Liu, and
  Liu}]{Song_Zhou_Fu_Liu_Liu_2021}
Wei Song, Shuhui Zhou, Ruiji Fu, Ting Liu, and Lizhen Liu. 2021.
\newblock \href {https://doi.org/10.18653/v1/2021.acl-long.327} {Verb metaphor
  detection via contextual relation learning}.
\newblock In \emph{Proceedings of the 59th Annual Meeting of the Association
  for Computational Linguistics and the 11th International Joint Conference on
  Natural Language Processing (Volume 1: Long Papers)}.

\bibitem[{Steen et~al.(2010)Steen, Dorst, Herrmann, Kaal, Krennmayr, and
  Pasma}]{Steen_Dorst_Herrmann_Kaal_Krennmayr_Pasma_2010}
Gerard~J. Steen, Aletta~G. Dorst, J.~Berenike Herrmann, Anna Kaal, Tina
  Krennmayr, and Trijntje Pasma. 2010.
\newblock \href {https://doi.org/10.1075/celcr.14} {A method for linguistic
  metaphor identification}.

\bibitem[{Su et~al.(2021)Su, Wu, and Chen}]{Su_Wu_Chen_2021}
Chang Su, Kechun Wu, and Yijiang Chen. 2021.
\newblock \href {https://doi.org/10.18653/v1/2021.findings-acl.109} {Enhanced
  metaphor detection via incorporation of external knowledge based on
  linguistic theories}.
\newblock In \emph{Findings of the Association for Computational Linguistics:
  ACL-IJCNLP 2021}.

\bibitem[{Su et~al.(2020)Su, Fukumoto, Huang, Li, Wang, and
  Chen}]{Su_Fukumoto_Huang_Li_Wang_Chen_2020}
Chuandong Su, Fumiyo Fukumoto, Xiaoxi Huang, Jiyi Li, Rongbo Wang, and Zhiqun
  Chen. 2020.
\newblock \href {https://doi.org/10.18653/v1/2020.figlang-1.4} {Deepmet: A
  reading comprehension paradigm for token-level metaphor detection}.
\newblock In \emph{Proceedings of the Second Workshop on Figurative Language
  Processing}.

\bibitem[{Tong et~al.(2021)Tong, Shutova, and Lewis}]{Tong_Shutova_Lewis_2021}
Xiaoyu Tong, Ekaterina Shutova, and Martha Lewis. 2021.
\newblock \href {https://doi.org/10.18653/v1/2021.naacl-main.372} {Recent
  advances in neural metaphor processing: A linguistic, cognitive and social
  perspective}.
\newblock In \emph{Proceedings of the 2021 Conference of the North American
  Chapter of the Association for Computational Linguistics: Human Language
  Technologies}.

\bibitem[{Tsvetkov et~al.(2013)Tsvetkov, Mukomel, and
  Gershman}]{Tsvetkov_Mukomel_Gershman_2013}
Yulia Tsvetkov, Elena Mukomel, and Anatole Gershman. 2013.
\newblock Cross-lingual metaphor detection using common semantic features.

\bibitem[{Wachowiak and Gromann(2023)}]{wachowiak2023does}
Lennart Wachowiak and Dagmar Gromann. 2023.
\newblock Does gpt-3 grasp metaphors? identifying metaphor mappings with
  generative language models.
\newblock In \emph{Proceedings of the 61st Annual Meeting of the Association
  for Computational Linguistics (Volume 1: Long Papers)}, pages 1018--1032.

\bibitem[{Wan et~al.(2021)Wan, Lin, Du, Shen, and
  Zhang}]{Wan_Lin_Du_Shen_Zhang_2021}
Hai Wan, Jinxia Lin, Jianfeng Du, Dawei Shen, and Manrong Zhang. 2021.
\newblock \href {https://doi.org/10.18653/v1/2021.findings-acl.173} {Enhancing
  metaphor detection by gloss-based interpretations}.
\newblock In \emph{Findings of the Association for Computational Linguistics:
  ACL-IJCNLP 2021}.

\bibitem[{Wang et~al.(2023)Wang, Li, Lin, Barrault, and
  Guerin}]{Wang_Li_Lin_Barrault_Guerin_2023}
Shun Wang, Yucheng Li, Chenghua Lin, Lo\"ic Barrault, and Frank Guerin. 2023.
\newblock Metaphor detection with effective context denoising.

\bibitem[{Wang et~al.(2021)Wang, Liu, Xu, Zhu, and Zeng}]{wang2021want}
Shuohang Wang, Yang Liu, Yichong Xu, Chenguang Zhu, and Michael Zeng. 2021.
\newblock Want to reduce labeling cost? gpt-3 can help.
\newblock \emph{arXiv preprint arXiv:2108.13487}.

\bibitem[{Wilks et~al.(2013)Wilks, Dalton, Allen, and
  Galescu}]{Wilks_Dalton_Allen_Galescu_2013}
Yorick Wilks, Adam Dalton, JamesF. Allen, and Lucian Galescu. 2013.
\newblock Automatic metaphor detection using large-scale lexical resources and
  conventional metaphor extraction.

\bibitem[{Ye et~al.(2022)Ye, Gao, Li, Xu, Feng, Wu, Yu, and
  Kong}]{ye2022zerogen}
Jiacheng Ye, Jiahui Gao, Qintong Li, Hang Xu, Jiangtao Feng, Zhiyong Wu, Tao
  Yu, and Lingpeng Kong. 2022.
\newblock Zerogen: Efficient zero-shot learning via dataset generation.
\newblock \emph{arXiv preprint arXiv:2202.07922}.

\bibitem[{Yoo et~al.(2021)Yoo, Park, Kang, Lee, and Park}]{yoo2021gpt3mix}
Kang~Min Yoo, Dongju Park, Jaewook Kang, Sang-Woo Lee, and Woomyeong Park.
  2021.
\newblock Gpt3mix: Leveraging large-scale language models for text
  augmentation.
\newblock \emph{arXiv preprint arXiv:2104.08826}.

\bibitem[{Zhang and Liu()}]{Zhang_Liu}
Shenglong Zhang and Ying Liu.
\newblock Metaphor detection via linguistics enhanced siamese network.

\bibitem[{Zhang and Liu(2023)}]{zhang2023adversarial}
Shenglong Zhang and Ying Liu. 2023.
\newblock Adversarial multi-task learning for end-to-end metaphor detection.
\newblock \emph{arXiv preprint arXiv:2305.16638}.

\end{thebibliography}
\bibliographystyle{acl_natbib}

\end{document}